\documentclass[conference]{IEEEtran}
\IEEEoverridecommandlockouts
\pdfoutput=1
\usepackage{cite}
\usepackage{amsmath,amssymb,amsfonts}
\usepackage{algorithmic}
\usepackage{graphicx}
\usepackage{textcomp}
\usepackage{xcolor}
\usepackage[mathscr]{eucal}
\usepackage[colorlinks,
            linkcolor=red,
            anchorcolor=blue,
            citecolor=green]{hyperref}  
\def\BibTeX{{\rm B\kern-.05em{\sc i\kern-.025em b}\kern-.08em
    T\kern-.1667em\lower.7ex\hbox{E}\kern-.125emX}}
\begin{document}

\title{Semantic Object-level Modeling for Robust Visual Camera Relocalization\\}

\author{
\IEEEauthorblockN{1\textsuperscript{st} Yifan Zhu}
\IEEEauthorblockA{\textit{School of Automation} \\
\textit{Beijing Institute of Technology}\\
Beijing, China \\
3120210915@bit.edu.cn}
\and
\IEEEauthorblockN{2\textsuperscript{nd} Lingjuan Miao}
\IEEEauthorblockA{\textit{School of Automation} \\
\textit{Beijing Institute of Technology}\\
Beijing, China \\
miaolingjuan@bit.edu.cn}
\and
\IEEEauthorblockN{3\textsuperscript{rd} Haitao Wu $^{*}$}
\IEEEauthorblockA{\textit{Aerospace Information Research Institute} \\
\textit{Chinese Academy of Sciences}\\
Beijing, China \\
wuht@aircas.ac.cn}
\and
\IEEEauthorblockN{4\textsuperscript{th} Zhiqiang Zhou}
\IEEEauthorblockA{\textit{School of Automation} \\
\textit{Beijing Institute of Technology}\\
Beijing, China \\
zhzhzhou@bit.edu.cn}
\and
\IEEEauthorblockN{5\textsuperscript{th} Weiyi Chen}
\IEEEauthorblockA{\textit{School of Automation} \\
\textit{Beijing Institute of Technology}\\
Beijing, China \\
3220195102@bit.edu.cn}
\and
\IEEEauthorblockN{6\textsuperscript{th} Longwen Wu}
\IEEEauthorblockA{\textit{School of Automation} \\
\textit{Beijing Institute of Technology}\\
Beijing, China \\
3120210908@bit.edu.cn}
}

\maketitle

\begin{abstract}
Visual relocalization is crucial for autonomous visual localization and navigation of mobile robotics. 
Due to the improvement of CNN-based object detection algorithm, 
the robustness of visual relocalization is greatly enhanced especially in viewpoints where classical methods fail.
However, ellipsoids (quadrics) generated by axis-aligned object detection 
may limit the accuracy of the object-level representation 
and degenerate the performance of visual relocalization system. 
In this paper, we propose a novel method of automatic object-level voxel modeling 
for accurate ellipsoidal representations of objects.
As for visual relocalization, we design a better pose optimization strategy for camera pose recovery, to fully utilize the projection characteristics of 2D fitted ellipses and the 3D accurate ellipsoids. All of these modules are entirely intergrated into visual SLAM system. 
Experimental results show that our semantic object-level mapping and object-based visual relocalization methods 
significantly enhance the performance of visual relocalization in terms of robustness to new viewpoints.
\end{abstract}


\section{Introduction}
Visual relocalization refers to the use of image sets, 3D point clouds, semantic objects, or other useful data to obtain the camera pose of query image frames, 
in order to solve the problem of camera uninitialization or localization failure 
in the Simultaneous Localization and Mapping (SLAM).
In practice, 6-DOF camera pose estimation based on known global scene maps has been widely applied in fields such as mobile robotics, autonomous driving, AR, and so on.   

A robust and effective visual relocalization algorithm is extremely important for visual SLAM systems. 
It is quite common for motion blur and rapid movement to cause visual SLAM tracking failure, 
which seriously affects the widespread application of visual SLAM systems. 
As a consequence, the robust operation of the visual relocalization module is essential for recovering the SLAM process. 
It is crucial to enhance the robustness (reliability) of the visual relocalization module in order to 
solve the problems of robot loss and kidnapped robots that often occur in the indoor scenes.
\begin{figure}[htbp]
    \centerline{\includegraphics[scale=1]{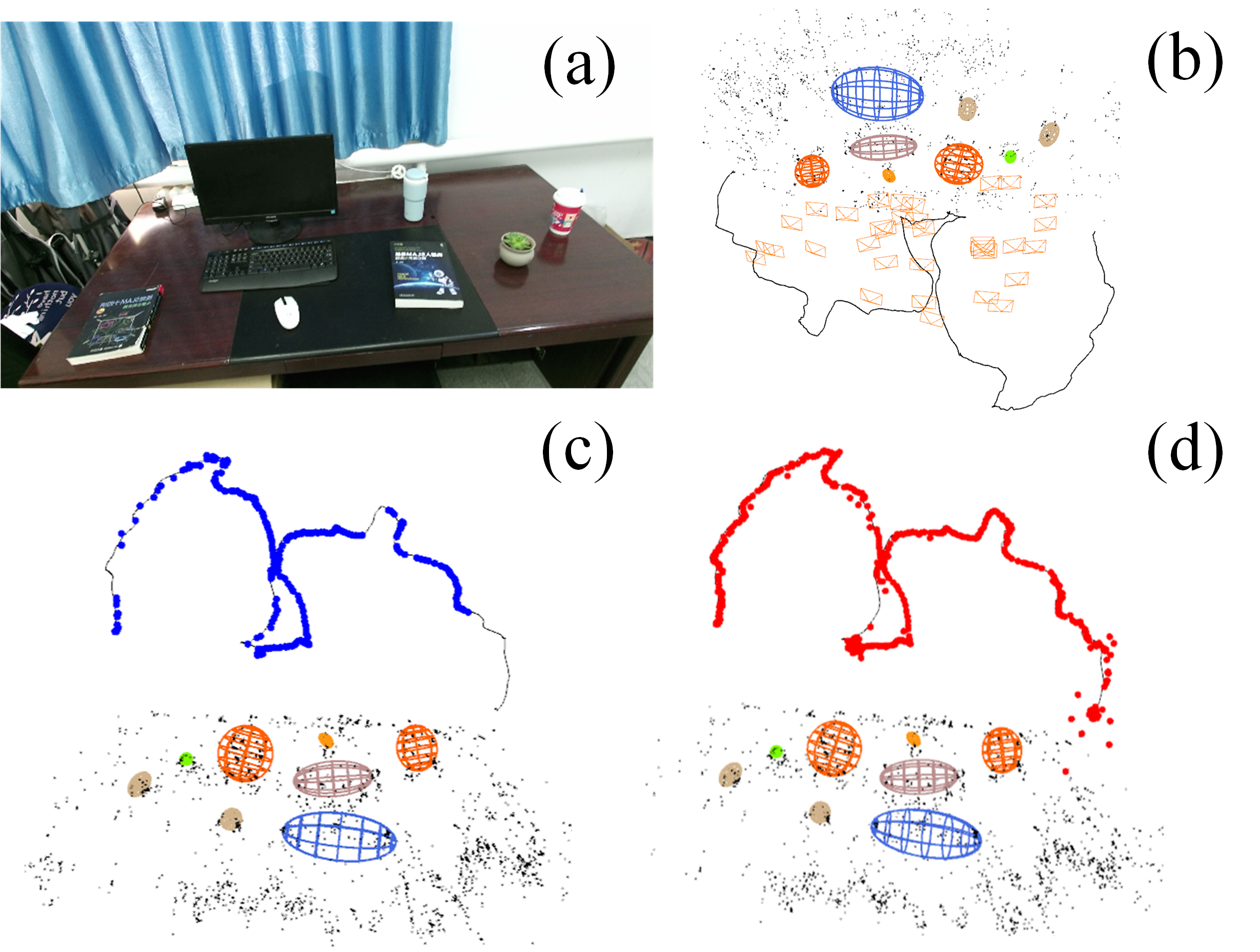}}
    \caption{Relocalization using known ellipsoid-based and point-based map. (a): A RGB frame in our collected video sequence for mapping. (b): There are some viewpoint changes between the yellow keyframes for mapping and black ground truth trajectory of relocalization sequence. (c)\&(d): Blue and red points are successfully relocalized frames using ORB-SLAM2\cite{ORB-SLAM2} and our method, repectively. Apparently, Our method allows the camera to be relocalized from viewpoints where ORB-SLAM2 fails.}
    \label{fig1}
\end{figure}

Current keypoint-based visual SLAM frameworks, such as ORB-SLAM2\cite{ORB-SLAM2}, depending on local ORB\cite{ORB} features and Bag-of-Words\cite{BoW} descriptors, search for the most similar reference image
in the image sets to obtain many matches between 2D key points in query images and 3D map points in the map. And through these matches, the PnP algorithm 
is then used to recover camera pose in a RANSAC loop. 
However, when there is a significant change of viewpoints between the query frame 
and the reference keyframe database, the local manual features of the image change significantly 
with the viewpoints, which makes it difficult to effectively perform feature matching, 
leading to camera relocalization failure.

However, due to the rapid improvement of CNN-based object detection, methods that use 3D geometric models (cuboids, ellipsoids) and 2D object detection to complete object-based camera pose estimation are emerging\cite{locUsingEllipsoids, RelocUsingEllipsoids1, QuadricSLAM, CubeSLAM, zinsDL, OA-SLAM}. In OA-SLAM\cite{OA-SLAM}, Zins et al. used object detection as constraints in the mapping process to automatically build object ellipsoidal representations on the fly, and utilized objects and points in the map for robust visual relocalization.

These methods use geometric models to roughly represent objects through 2D axis-aligned bounding box constraints, which have achieved desirable results in 2D-3D object-level association and camera pose estimation. But owing to the direct use of 2D centers of axis-aligned bounding boxes and corresponding central points in 3D ellipsoids, the PnP algorithm can only calculate camera poses that are not accurate enough, especially when the 3D ellipsoid representation is not accurate or the center point of the 2D detection box is not coherent with the projection of center point in the 3D ellipsoid. In \cite{zinsDL}, Zins et al. proposed a learning-based method which detects improved elliptic approximations of 2D detected objects which are coherent with the 3D ellipsoids in terms of perspective projection. It shows remarkable results but needs manual annotations again when encountering new scenes, which means hard to integrate into SLAM system.

In order to enhance the performance of relocalization in visual SLAM in terms of robustness while ensuring accuarcy, we explore a novel way to obtain accurate ellipsoidal representations (with accurate poses) for static object landmarks in unknown indoor scenes. As for relocalization, when a query image arrives, 
instance segmentation masks on query images are used to calculate fitted ellipses for regular objects, and a object-based optimization strategy is imposed to refine the initial pose computed by PnP algorithm. Our main contributions are as follows:

\begin{itemize}
\item We propose a novel mapping method to obtain accurate ellipsoidal representations of semantic object-level landmarks, leveraging object-level voxel modelling and automatic object-level associations.
\item We design a object-based relocalization strategy to fully use projection characteristics of 2D fitted ellipses
and the built 3D accurate ellipsoids.
\item Our object-level mapping method and object-based relocalization strategy are entirely intergrated into RGBD SLAM system, robust to a variety of viewpoints, adaptive to unknown indoor scenes, and running in real-time.
\end{itemize}

\section{Related works}
\subsection{Semantic Object-level Mapping}

To autonomously navigate in real-world environments, robots must be able to perceive complex and unstructured scenes, and build object-oriented scene maps.

Recent methods \cite{VolumetricInstanceMapping, PointBasedInstanceMapping, IncrementalObjectDatabase, Loop} used  geometric representations like point, mesh, voxel or TSDF model to build semantically meaningful, object-level entities with fine grain. 

Some other recent methods \cite{locUsingEllipsoids, RelocUsingEllipsoids1, QuadricSLAM, CubeSLAM, zinsDL, OA-SLAM, Loop} chose to build coarse-grained geometric representations (cuboids or ellipsoids) of objects to enhance visual SLAM or modules such as loop closing and relocalization in visual SLAM. In \cite{Loop}, Lin et al. relied on 2D instance segmentation mask to extract object-level voxel models, and used these voxel models to calculate 3D cuboids for loop closing in visual SLAM. Similarly, in CubeSLAM\cite{CubeSLAM}, Yang et al. used 3D cuboids to represent objects in the map. These cuboids are jointly optimized with camera poses and landmarks. In addition to representing objects with 3D cuboids, there are also some methods that use ellipsoids (quadrics) to represent objects in the map. In QuadricSLAM\cite{QuadricSLAM}, Nicholson et al. derived a SLAM formulation that uses dual quadrics (ellipsoid) as 3D landmarks to represent objects. OA-SLAM\cite{OA-SLAM}, proposed by Zins et al, also uses ellipsoidal representations for objects during mapping process. It is noteworthy that these coarse ellipsoidal representations use 2D axis-aligned bounding boxes for constraints and optimization, and can only roughly represent the pose of objects.

\subsection{Object-based Camera Pose Estimation}
Recent works of using objects to calculate camera pose in visual SLAM can be divided into two types. The first is to couple the object as a landmark with the SLAM system, and optimize the camera pose and object landmark jointly during the SLAM process\cite{CubeSLAM, QuadricSLAM}. The second type of object landmark has a shallow coupling with SLAM, only using the camera poses provided by SLAM to extract objects in the image. They build objects in the mapping process, and then use these objects to enhance SLAM modules such as loop closing and visual relocalization\cite{Loop, RelocUsingEllipsoids1, OA-SLAM}. They take full use of current remarkable object detection or instance segmentation algorithm like Faster R-CNN\cite{FasterRCNN}, Mask R-CNN\cite{MaskRCNN} or YOLOs\cite{YOLOv8}. In OA-SLAM\cite{OA-SLAM}, Zins et al. used objects with high-level semantics and points with better spatial localization accuarcy to improve visual relocalization module, which often fails in ORB-SLAM2\cite{ORB-SLAM2} when there are a variety of viewpoints. Though promising, we go one step futher and propose a new method to generate more accurate ellipsoidal models for objects in the map. By using these accurate ellipsoidal models, we design a strategy when relocalizing the query images in order to futher enhance the robustness of relocalization when encountering big view changes.

\section{Method}
\subsection{System Overview}
\begin{figure}[htbp]
    \centerline{\includegraphics[scale=0.65]{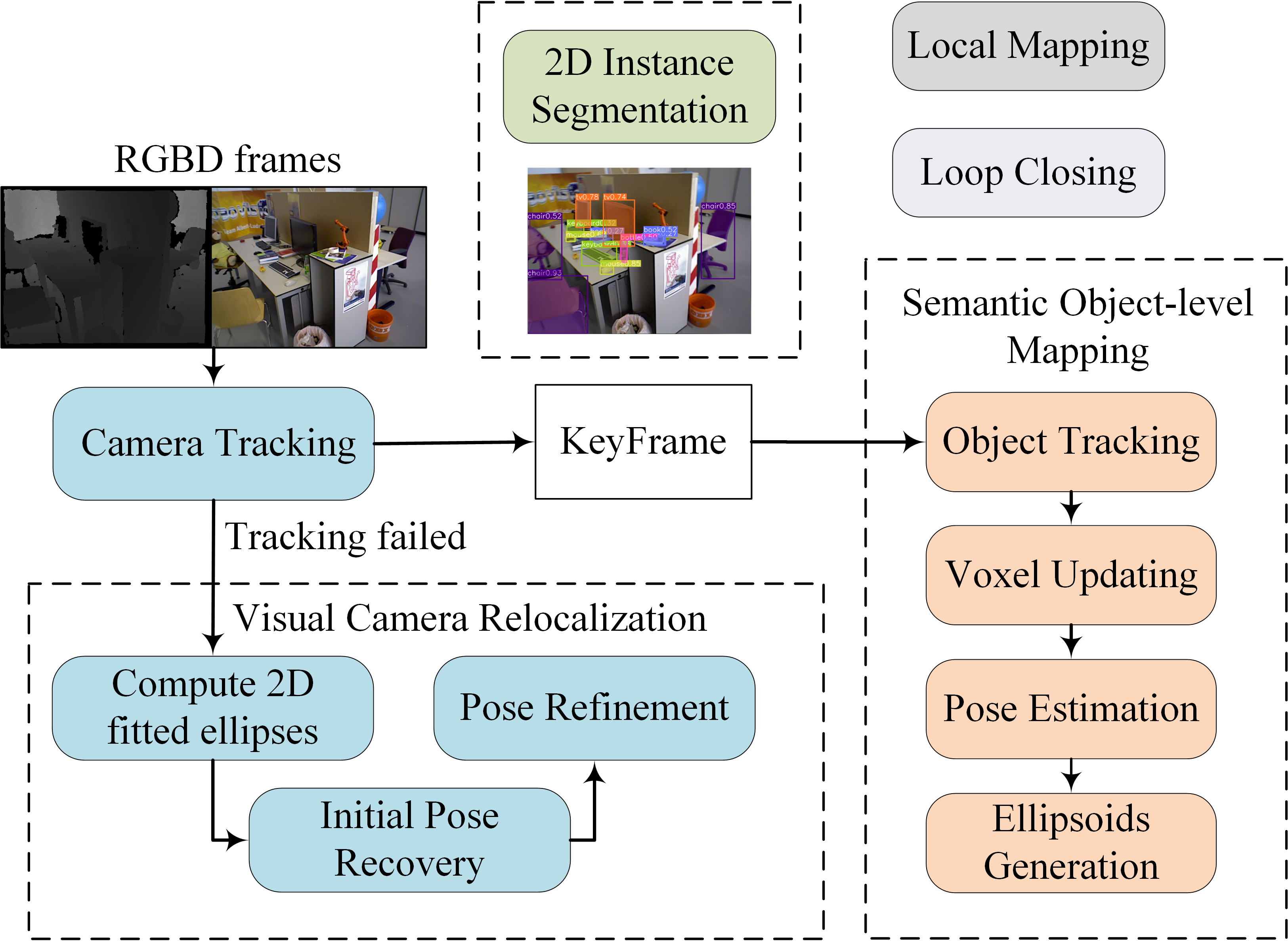}}
    \caption{System overview: dashed boxes are newly added elements within ORB-SLAM2 backbone.The modules filled with different colors are run in separate thread.}
    \label{fig2}
\end{figure}
Our proposed framework is detailed in Fig. \ref{fig2}. It is based on ORB-SLAM2 backbone and consists of two main parts (semantic object-level mapping and visual camera relocalization). Keyframes provided during the camera tracking process are used for semantic object-level mapping. Among them, the keyframes are processed by the instance segmentation thread to obtain bounding boxes and instance segmentation masks. We use the camera pose, RGBD data, and instance segmentation data of keyframes for mapping process. As for relocalization module, the instance segmentation is similarly used for RGB query frame in order to get fitted 2D ellipses for detected regular objects. Then the 2D ellipses are associated with 3D ellipsoids built in mapping process to estimate and recover camera pose in case of SLAM tracking lost.

\subsection{Object-level Voxel Modelling}\label{voxelmodelling}
In order to accurately describe the geometric appearance structure of objects in indoor scenes, facilitate object tracking (described in section \ref{objs_tracking}) and accurate ellipsoid generation (described in section \ref{acc_ell_gen}), we use voxel models to process the original object segmentation data, in order to achieve accurate 3D object extraction. We perform 2D instance segmentation\cite{YOLOv8} to obtain 2D observation of each object on keyframes determined by camera tracking in ORB-SLAM2.

According to the poses, instance segmentation masks and depth maps of keyframes, we can obtain dense point clouds of objects at each keyframe perspective. In practical situations, due to the limitation of segmentation mask errors, dense point clouds of actual objects contain a large amount of noise and useless background information, which will seriously deteriorate the accuracy of 3D object entities.

Hence, similar to the method in \cite{Loop}, we use semantic label probability values to replace the occupancy probability of the grid map. Semantic label probability stored in each voxel of object entities is dynamiclly updated according to continuous observations from RGBD sequence as introduced in \cite{Octomap}. This method is experimentally demonstrated that it's effective and efficient to filter noise produced by segmentation masks and sensors.

Assuming that there are $M$ categories of objects in the scene, for the $i$-th object, its semantic category ID is $c_i\in\{1,2,\dots,M\}$. The number of nodes used to represent the probability voxel model of the $i$-th object is $N_i$, where the node number in $i$-th object is denoted $n\in\{1,2,\dots, N_i\}$. And the semantic label probability of $c_i$ for the $n$-th node of $i$-th object is denoted $P(c_i, n)$. After we get the observations in frame $T$, if a prior probability $P(c_i,n|z_{1:T-1})$ is known from the first frame to the $(T-1)$-th frame, then semantic label probability $P(c_i,n|z_{1:T})$ will be updated according to the following formulation: 
\begin{equation}\label{eq1}
\resizebox{0.9\hsize}{!}{$
        \begin{aligned}
            &P(c_i,n|z_{1:T}) \\
            &=[1+\frac{1-P(c_i,n|z_T)}{P(c_i,n|z_T)}\frac{1-P(c_i,n|z_{1:T-1})}{P(c_i,n|z_{1:T-1})}\frac{P(c_i,n)}{1-P(c_i,n)}]^{-1}_{.}
        \end{aligned}
$}
\end{equation}

When the object is tracked by objects tracking (described in \ref{objs_tracking}) by $\omega_0$ times, voxel model of the objects are inserted into the global map, and the label with the maximum semantic probability is selected as the object's final semantic label. Then, according to the preset filtering threshold, the voxel whose semantic label probability less than the threshold is cleared stage by stage. $\omega_0$ and initial semantic label probability is set to 3 and 0.5.

\begin{figure*}[htbp]
    \centerline{\includegraphics[scale=1]{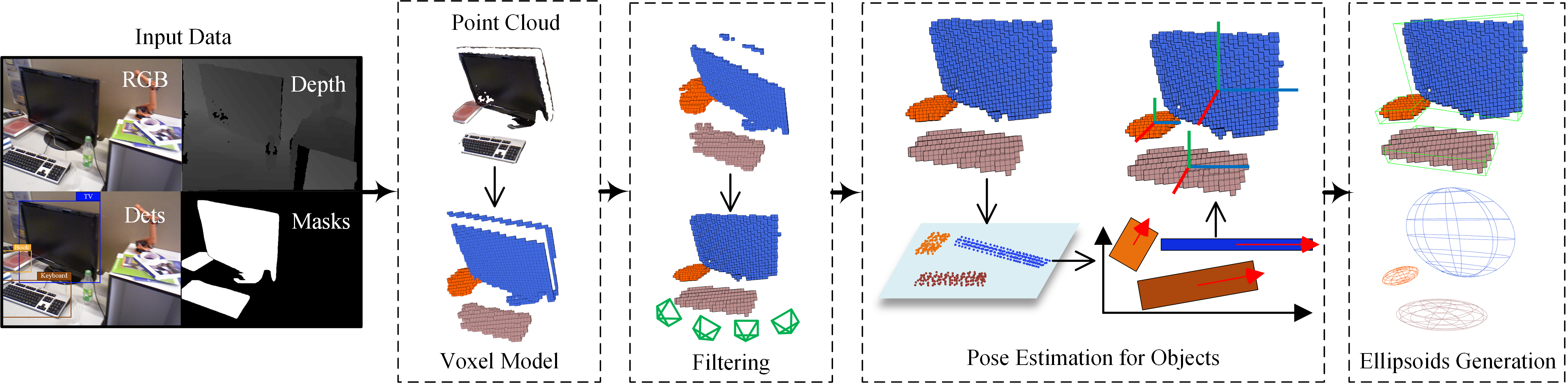}}
    \caption{Overview of object-based Mapping pipeline. TV monitor, keyboard and book are three different sample objects. Object Tracking procedure is not shown in the figure. The associated 3D object is continuously updated and filtered through the 2D images, and the pose and ellipsoid model are updated simultaneously.}
    \label{fig3}
\end{figure*}

\subsection{Objects Tracking}\label{objs_tracking}
Objects tracking is divided into two categories: 2D objects tracking between adjacent frames and 3D objects matching. 2D object tracking is to distinguish whether the observation of adjacent frames for the same class of objects is the same object or not; 3D object matching is to determine whether the new object is an existing object in the scene or an unknown new object. When the same object is observed in continuous frames, it is mainly the 2D object tracking that takes effect; when the object disappears from the field of view for a period of time, it then suddenly appears (such as a loop), then 3D object matching is in effect.

\textbf{2D objects tracking}. Similarly to \cite{DeepSort,OA-SLAM}, 2D multiple objects tracking is performed between adjacent keyframes. Because 2D-3D object correspondences needed to be transferred from previous $(t-1)$-th frame to the current $t$-th frame through 2D objects tracking. The voxels of an object $O_{t-1,i}$ (where subscript denote $i$-th object in the map detected in frame $t-1$) are projected onto camera frame to form 2D point set, which generates known 2D bounding boxes with axis alignment. The optimal associations between the known 2D bounding boxes and the current 2D detected bounding boxes are completed using the Hungarian algorithm\cite{KM}. Thus, each element in cost matrix is defined as:
\begin{equation}
    \resizebox{0.9\hsize}{!}{$
        \begin{split}
            \begin{aligned}
                cost(i,j)=&\lambda_1f(O_{t-1,i}, o_{t, j})+ \\
                &\lambda_2(1-\text{IoU}(\text{bbox}(\text{proj}(O_{t-1,i})), \text{bbox}(o_{t,j}))),\label{eq2}
            \end{aligned}
        \end{split}
    $}
\end{equation}
\begin{equation}
    \label{eq3}
    f(a,b)=\left\{
    \begin{aligned}
        \ &1, \quad \text{if a and b have same semantic ID,} \\
        \ &0, \quad \text{otherwise,}
    \end{aligned}
    \right.
\end{equation}
where $\lambda_1$ and $\lambda_2$ ($\lambda_1\gg\lambda_2$) are the weights of semantic cost term and reprojection cost term, respectively. Besides, operations such as \textit{proj}, \textit{bbox} and \textit{IoU} means projecting 3D object to 2D point set, calculating bounding box for 2D point set and computing intersection-over-union (IoU), repectively. For the i-th object $O_{t-1,i}$ tracked by last frame and the j-th 2D detection $o_{t,j}$ in the current frame, the semantic label ID consistency must be met first, and secondly, the minimum reprojection error based on IoU is required to be considered as a 2D-3D matching pair.

\textbf{3D objects matching}. For newly detected 2D objects that are not matched in the current frame $t$ in 2D objects tracking, we need to determine whether their corresponding 3D objects already exist in the map. We generate a voxel model $v$ for a newly detected 2D object and then use voxel association to determine whether the detected object already exists in the map. Specifically, we traverse the map objects and retain the object $v_i$ that matches the semantic label ID. Firstly, set the search radius r, search and record the number of voxels in $v_i$ that are closest to the voxels in $v$, denoted as $n(v, v_i)$. Similarly, $n(v_i, v)$ can be calculated in the same way. The successful judgment of 3D object matching is as follows:
\begin{equation}
    \label{eq4}
    \left\{
    \begin{aligned}
        \ n(v, v_i)>\xi\text{min}(v, v_i),\\
        \ n(v_i, v)>\xi\text{min}(v, v_i),
    \end{aligned}
    \right.
\end{equation}
where $\text{min}(v, v_i)$ means minimum nodes for each of them. $\xi$ is a threshold set to 0.5 in our experiment.

\subsection{Accurate Ellipsoids Generation}\label{acc_ell_gen}
We use ellipsoids(dual quadratics) to represent individual objects in the map. Similary to \cite{QuadricSLAM, OA-SLAM}, An ellipsoid is defined by nine parameters including rotation, position and three semi-axes. However, compared with these methods, ellipsoids generated by our method has better accuracy in terms of the above nine parameters. In order to obtain the accurate ellipsoid $\mathbf{Q^*}$ for an inserted object in the map, center coordinates $t=[x,y,z]^T$ and rotation $\mathbf{R}(\theta)$ relative to the world coordinate system, size $(h, w, l)$ of the voxel model are calculated. Ellipsoid $\mathbf{Q^*}$ is determined by transformation $Z$ and initial ellipsoid $\mathbf{\Breve{Q}^*}$ centred at the origin.
\begin{equation} \label{eq5}
    \mathbf{Q^*}=
    \mathbf{Z}^\top
    \mathbf{\Breve{Q}^*}
    \mathbf{Z}, 
\end{equation}

\begin{equation}\label{eq6}
    \!\!
    \mathbf{Z} = 
    \begin{pmatrix}
        \mathbf{R}(\theta) & t^\top \\
        \mathbf{0}^\top_{3} & 1
    \end{pmatrix}
    \,\text{and}\,
    \mathbf{\Breve{Q}^*}=
    \text{diag}(\frac{l^2}{4},\frac{w^2}{4},\frac{h^2}{4},-1).
    \!\!
\end{equation}

\textbf{Pose estimation for objects.} We assume that all objects are placed on the ground and we have calculated the center coordinates of the objects, so the six-DOF pose estimation is reduced to one (yaw angle). 3D voxels of a object are projected onto the ground, and the PCA algorithm is implemented to compute the red main axis (shown in \ref{fig3}) of the projected 2D points set. The yaw angle of an object is determined by the angle of the computed main axis. Hence, we can obtain the accurate 9-DOF ellipsoidal representations which actually are inscribed inside the cuboids generated by the computed pose for all the objects in the map. It is worth mentioning that compared to our method, works like\cite{QuadricSLAM,OA-SLAM} merely use axis-aligned bounding box constraints to generate coarse ellipsoids, resulting in lower accuracy in object representation.

\begin{figure*}[htbp]
    \centerline{\includegraphics[scale=1]{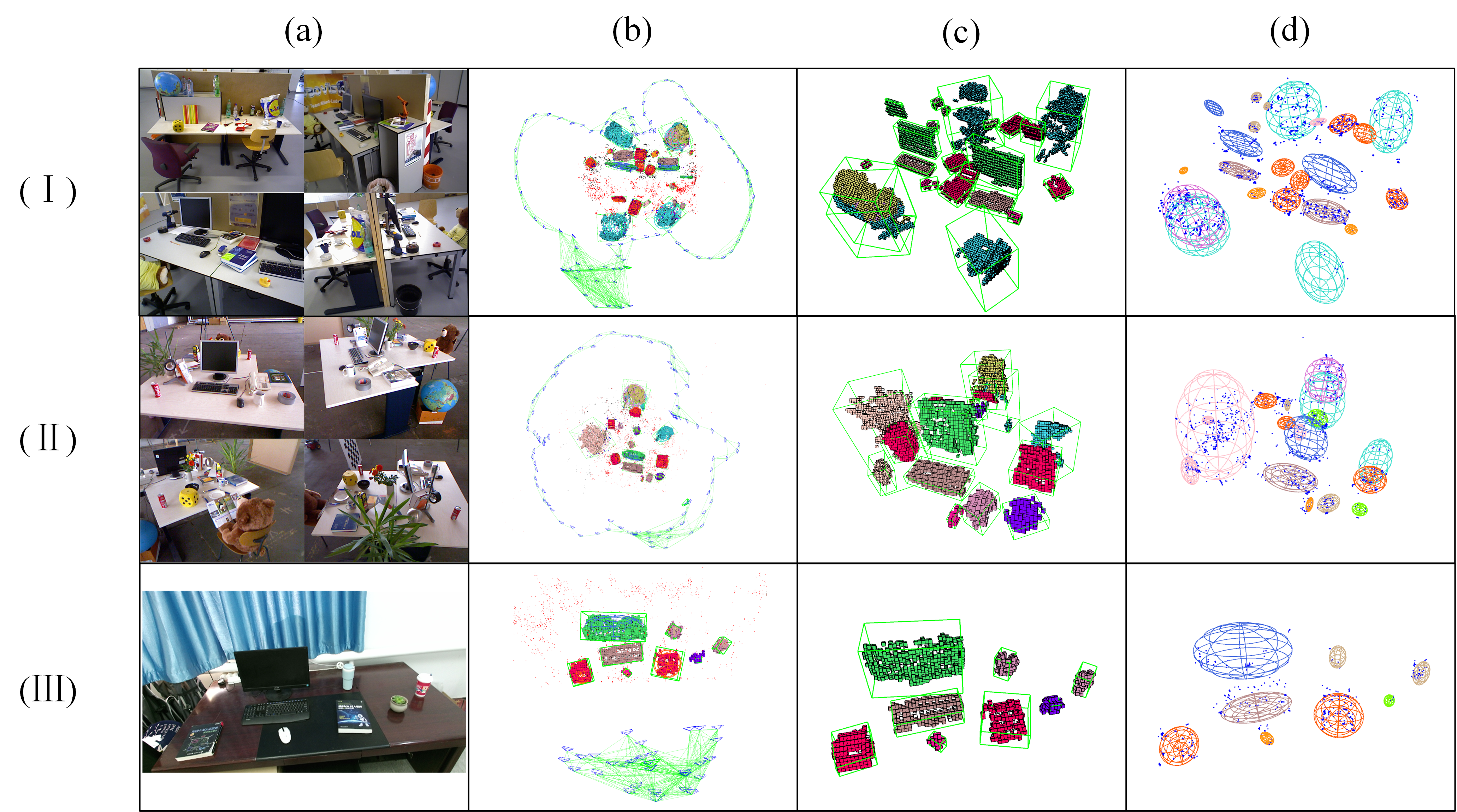}}
    \caption{Accurate ellipsoids mapping in three different scenes.((\uppercase\expandafter{\romannumeral1}, \uppercase\expandafter{\romannumeral2}, \uppercase\expandafter{\romannumeral3}) are TUM \textit{fr3/long\_office\_household}, TUM \textit{fr2/desk} and customized scene, respectively.) From left to right, the columns are the RGB(s) of three scenes, mapping process, voxel and cuboids, ellipsoid model and ORB points on objects surface.}
    \label{fig4}
\end{figure*}

\subsection{Object-based Relocalization}\label{reloc}
The main precedure of object-based relocalization is shown in \ref{fig2}. When camera tracking fails and the query frame needs to search for its own pose based on the already built global map, the object-based relocalization module is enabled. RGB, instance segmentation masks and axis-aligned bounding boxes of query image are sent to the relocalization module. It is noted that objects with projection masks presented as simple convex polygons are defined as regular objects. For regular objects, we fit the observation ellipses based on a 2D masks; For other objects, inscribed observation ellipses are obtained based on the axis-aligned bounding boxes. After obtaining the fitted observation ellipse, initial pose recovery and pose refinement for query image will be discussed seperately.

\textbf{Initial pose recovery.} An ellipsoid projects as an ellipse under any viewpoints, whose equation can be expressed in a closed-form manner using the dual space\cite{OA-SLAM}. In that space, for a 2D observation ellipse $\mathbf{C^*}$ and corresponding 3D ellipsoid $\mathbf{Q^*}$, if the projection matrix is $\mathbf{P}$, there is the following projection equation\cite{MVGbook}:
\begin{equation}\label{eq7}
    \mathbf{C^*} = \mathbf{P}\mathbf{Q^*}\mathbf{P}^\top.
\end{equation}

Therefore, initial pose recovery problem involves finding the matching relationship between the observation ellipses of query frame and the ellipsoids in the map, and solving for projection matrix of the query frame. We use the method (P3P loop) introduced in \cite{OA-SLAM, zinsDL} to jointly determine object correspondences between the query image and the map, and estimates the initial pose of the camera. Compared to them, we choose different fitting methods based on whether the 2D mask is regular or not. If boundary of the mask is roughly a quadrilateral, the object is considered as regular object and 2D point set in the mask is used to fit the observation ellipse; Otherwise, the ellipse that is inscribed on the bounding box is directly used as the observation ellipse. Obviously, observation ellipse fitted by mask for regular object is more in line with the projection characteristics of an ellipse-ellipsoid pair.

\textbf{Pose refinement.} Refinement of initial camera pose uses accurate ellipsoids $\mathbf{Q^*}$ in the map and observation ellipses set denoted $\chi$. Coarse pose of the query image and 2D-3D object correspondences $f(\cdot)$ are jointly obtained by the P3P loop. The robust kernel function $\rho(\cdot)$ and the covariance matrix $\mathbf{\Sigma}$ set based on the elliptical area are applied to the following formula:
\begin{equation}\label{eq8}
    \{\mathbf{R^*},t^*\}=\underset{\mathbf{R},t}{\text{argmin}}\underset{i\in\chi}{\Sigma}\rho(\mathscr{W}_2^2(\mathbf{E}_i, \mathbf{P}\mathbf{Q^*}_{f(i)}\mathbf{P}^\top)_{\mathbf{\Sigma}}),
\end{equation}
where $\mathscr{W}_2^2$ is the Wasserstein distance between two Gaussian distributions which are determined by observation ellipse $\mathbf{E}_i$ and projected ellipse based on camera projection matrix $\mathbf{P}$. In OA-SLAM\cite{OA-SLAM}, Zin et al. detailed the process but for ellipsoidal optimization in map. The object-based camera pose refinement strategy prevent the camera pose from deteriorating due to object occlusion, inaccuray of the observation ellipses, and other reasons. It is because the accuracy of ellipsoidal representations for objects in the map, and observed 2D fitted ellipses in line with the projection characteristics, we make the pose refinement strategy better use of ellipse-ellipsoid projection equation \ref{eq7} in the pinhole camera model. By the way, the points produced by ORB-SLAM2 also can be used to further refine camera pose like OA-SLAM.


\section{Experiments}
\subsection{Experimental Settings}
To evaluate our object-based relocalization method and mapping method, we used three common indoor scenes including TUM RGBD datasets\cite{TUMdataset} (\textit{fr3/long\_office\_household} and \textit{fr2/desk}) and one customized RGBD dataset collected in our office. In addition to public benchmark, we used Kinect v2 RGBD camera to capture the RGBD sequences at 640$\times$360 resolution with 30 frames per second.

It is noteworthy that each scene contains two video sequences. The first sequence is used for mapping, and the other sequence with a significant difference in perspective from the mapping is used for validating the relocalization algorithm. The ground truth poses of relocalized frames in the second video sequences are obtained using state-of-the-art visual SLAM method\cite{ORB-SLAM2} with loop correction if needed.

\begin{table*}[htbp]
    \caption{Evaluations of Relocalization Performance in Three Different Scenarios}
    \begin{center}
    \begin{tabular}{|c|c c c|c c c|c c c|}
    \hline
    &\multicolumn{3}{|c|}{\textbf{TUM \textit{fr3/long\_office\_household}}$^\mathrm{1}$}&\multicolumn{3}{|c|}{\textbf{TUM \textit{fr2/desk}}$^\mathrm{1}$}&\multicolumn{3}{|c|}{\textbf{customized}$^\mathrm{2}$} \\
    \textbf{Methods} & pos.err.(cm)&rot.err.($^{\circ}$)& valid(\%)&pos.err. (cm)&rot.err.($^{\circ}$)&valid(\%)&pos.err.(cm)&rot.err.($^{\circ}$)&valid(\%) \\ 
    \hline
    ORB-SLAM2\cite{ORB-SLAM2}(points)& \textbf{4.34} & \textbf{1.48} & 60.44 & \textbf{3.44} & \textbf{0.49} & 17.85 & \textbf{2.64} & \textbf{0.98} & 47.61  \\
    OA-SLAM\cite{OA-SLAM}(objects)& 17.60 & 7.06 & 42.73 & 20.29 & 9.03 & 45.49 & 11.35 & 8.12 & 87.61 \\
    OA-SLAM(objects+points)& 7.06 & 3.68 & 45.94 & 9.50 & 5.67 & 56.89 & 4.69 & 2.36 & 95.75 \\
    Ours(objects) & 11.15 & 4.92 & 56.31 & 16.53 & 6.17 & 60.49 & 9.87 & 6.44 & 93.98 \\
    Ours(objects+points) & 5.16 & 2.40 & \textbf{60.68} & 7.58 & 3.37 & \textbf{69.33} & 2.96 & 1.19 & \textbf{96.28} \\
    \hline
    \multicolumn{7}{l}{$^{\mathrm{1}}$OA-SLAM and our method use different maps built seperately.} \\
    \multicolumn{7}{l}{$^{\mathrm{2}}$OA-SLAM and our method use the same map built by our method.}
    \end{tabular}
    \label{tab0}
    \end{center}
\end{table*}

Additionally, the object detector or segmentation method used in the front-end directly affects performance of the semantic object-level mapping. We used YOLO v8\cite{YOLOv8} pre-trained in COCO datasets\cite{COCO} as our detector and instance segmentation algorithm without any fine-tuning. In order to ensure that the object landmark categories used in the mapping are consistent, OA-SLAM\cite{OA-SLAM} also used the same pre-trained YOLO v8 detector and disabled some objects.

\subsection{Semantic Object-level Mapping}
Our semantic object-level mapping method was evaluated on three common indoor scenes demonstrated in Fig. \ref{fig4}. In this demonstration, we can see that the voxel models of the common objects (TV monitors, chairs, books, keyboards, teddy bear, etc) are clearly generated by using our object-level voxel modelling method (\ref{voxelmodelling}). Simultaneously, pose of each voxel model is computed (\ref{acc_ell_gen}) and represented by rotated 3D cuboid. Ultimately, accurate ellipsoidal representations for all objects in the map are generated for visual camera relocalization if SLAM tracking fails. This semantic object-level mapping algorithm which is entirely integrated into ORB-SLAM2 can run in real-time and automatically in a seperate thread.

\begin{figure}[htbp]
    \centerline{\includegraphics[scale=0.85]{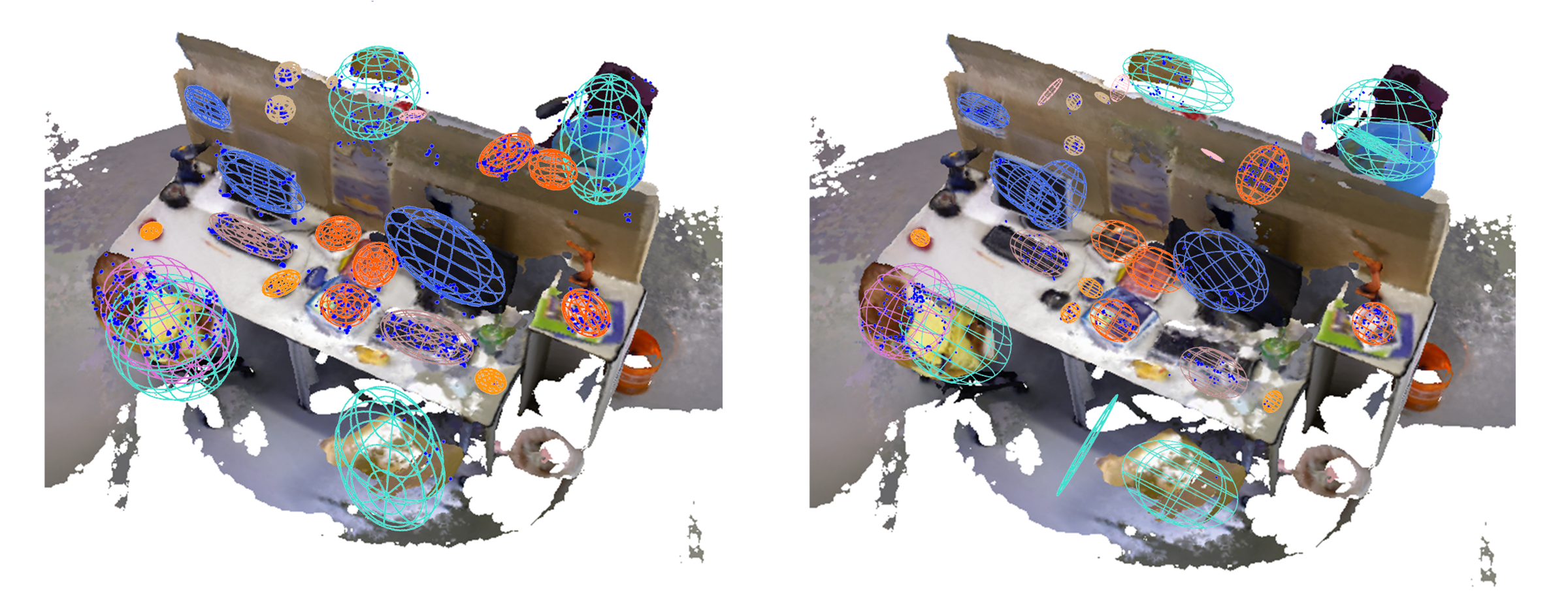}}
    \caption{Mapping comparision: Our Method and OA-SLAM\cite{OA-SLAM} on ellipsoidal landmarks mapping on TUM \textit{fr3/long\_office\_household} show that our method has better accuracy in ellipsoidal representations.}
    \label{fig5}
\end{figure}

Fig. \ref{fig5}. shows the object mapping comparision between our method and OA-SLAM\cite{OA-SLAM}. It is apparent that our method has better accuracy in ellipsoidal representation, which is beneficial to the enhancement of accuracy and robustness of our object-based relocalization method (\ref{reloc}).

\begin{figure}[htbp]
    \centerline{\includegraphics[scale=0.30]{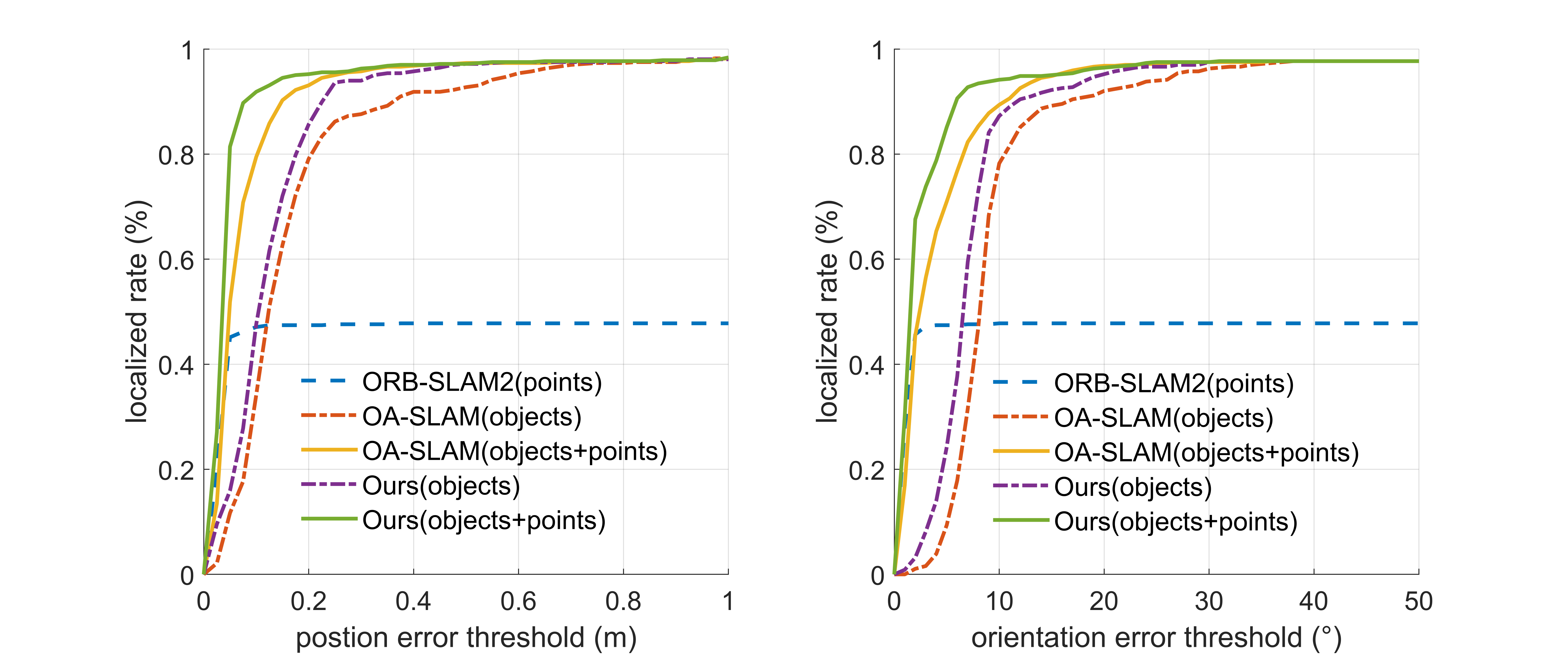}}
    \caption{Percentage of estimated camera positions and orientations with respect to corresponding error thresholds evaluated on customized dataset. Our method and OA-SLAM use the same ellipsoidal map built by our method. It shows that our relocalization strategy can take full use of the accurate 3D ellipsoidal landmarks and achieve optimal performance.}
    \label{fig6}
\end{figure}

\begin{figure}[htbp]
    \centerline{\includegraphics[scale=0.85]{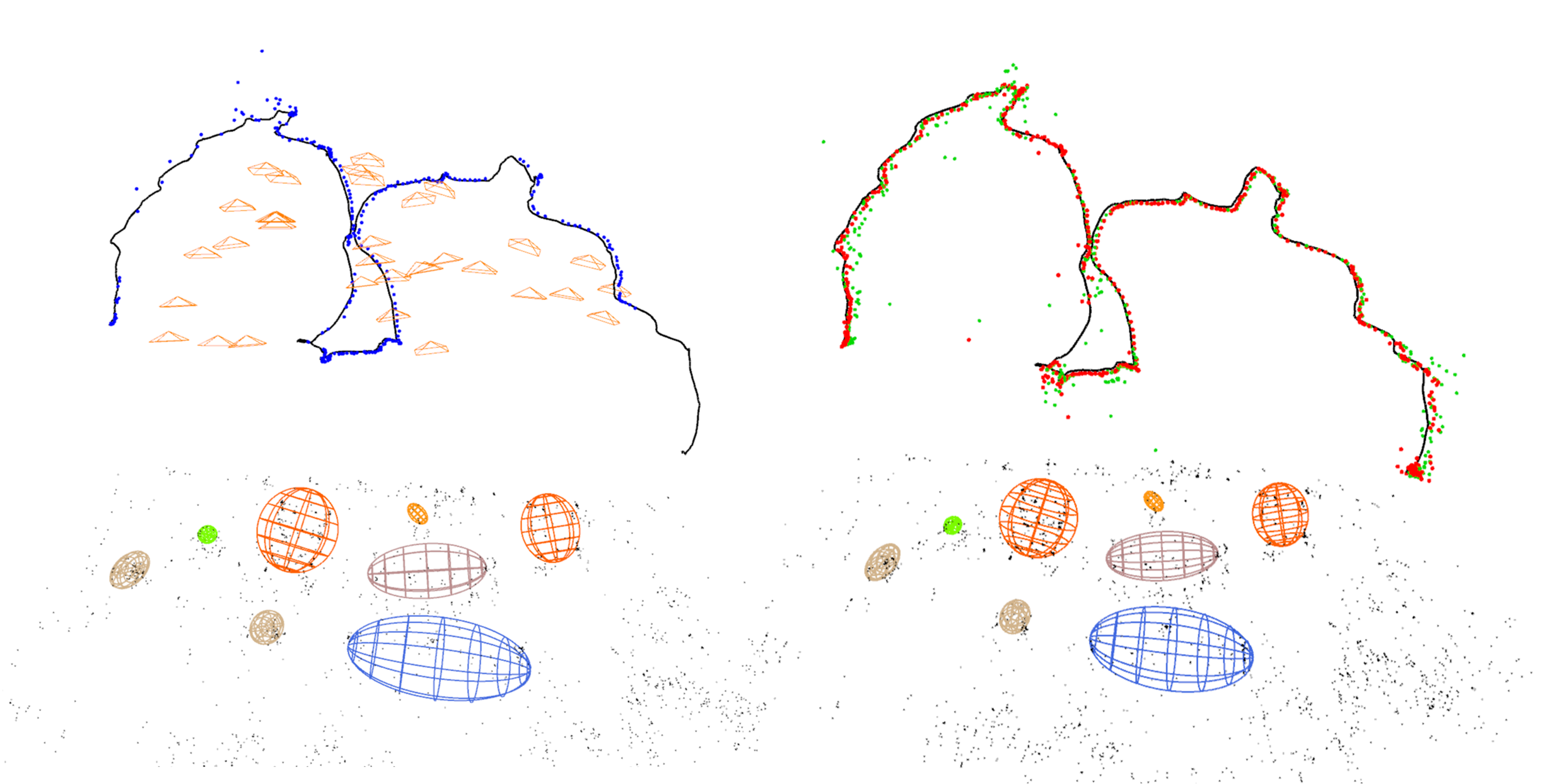}}
    \caption{Comparision of relocalization trajectories using different maps built seperately on customized dataset. Note that the map built by OA-SLAM is not shown in this figure. Yellow keyframes are used for mapping and have some viewpoint changes compared to black ground truth trajectories. Blue, green and red relocalization points are ORB-SLAM2(reloc), OA-SLAM(reloc) and our method, respectively. It shows that our valid relocalized query frames denoted as red points focus more on black ground truth.}
    \label{fig7}
\end{figure}

\subsection{Visual Relocalization}
The evaluations of our visual relocalization algorithm were conducted on the second different sequences with different viewpoints in pre-built three scenes. We used ORB-SLAM2\cite{ORB-SLAM2} and OA-SLAM\cite{OA-SLAM} as our comparision models. The map used by ORB-SLAM2 is the ORB point cloud map built using the first sequence of each scene. Besides, there are two types of maps used in OA-SLAM and our method. The first is mere object-level maps with ellipsoid representations, and the second is objects plus points maps, all built on the first sequence of each scene.

TABLE \ref{tab0} shows the quantitative evaluation results of relocalization performance. We evaluated the accuracy and robustness of each method by using the median error of position (less than 30 cm) and rotation (less than 30 degrees), as well as the proportion of query frames that successfully relocalized while satisfying both position and rotation thresholds in the second sequences. It shows that our method can achieve higher valid ratio and accuarcy in visual relocalization than OA-SLAM thanks to our semantic object-level mapping method and pose refinement strategy. Apart from that, Fig. \ref{fig7} is the visual display of comparisons between ground truth trajectory and relocalized frames represented by points if successfully. Therefore, according to TABLE \ref{tab0}, Fig. \ref{fig6} and Fig. \ref{fig7}, when using only objects for localization, both our mapping method and relocalization strategy have a promoting effect on the relocalization performance. If objects plus points are used together, the robustness of visual relocalization is greatly improved when the overall accuracy is close to ORB-SLAM2.

\section{Conlusions}
In this conference paper, we propose a novel semantic object-level mapping method and object-based visual camera relocalization strategy, all of which are totally integrated into ORB-SLAM2 backbone. Rather than generating ellipsoid representations using bounding box constraints, we use voxels to model object entities and directly computes more accurate ellipsoid representations, in order to better represent the position and pose of the objects in unknown indoor scenes.
Due to the full use of accurate ellipsoid representations built in the proposed mapping process, we can make the relocalization of visual SLAM more robust to large viewpoint changes while ensuring accuracy. 

\bibliography{References}

\end{document}